\newif\ifanonymousversion
\anonymousversionfalse

\ifanonymousversion
  \documentclass[sigconf]{acmart}
  \setcopyright{none}
  \renewcommand\footnotetextcopyrightpermission[1]{}
\else
  \documentclass[sigconf]{acmart}
  \setcopyright{none}
   \renewcommand\footnotetextcopyrightpermission[1]{}
\fi

\usepackage{caption}
\usepackage{subcaption}
\usepackage{graphicx}
\usepackage{xcolor}
\usepackage{soul}
\usepackage{url}
\usepackage{tabularx}
\usepackage{multirow}
\usepackage{multicol}
\usepackage{adjustbox}
\usepackage{hyperref}
\usepackage{array}
\usepackage{algorithm}
\usepackage{algpseudocode}
\usepackage{textcomp}
\usepackage{amsmath,amsfonts,amsthm}

\newcommand{\nsf}[1]{\href{https://www.nsf.gov/awardsearch/showAward?AWD_ID=#1}{#1}}

\begin{document}

%%
%% Title
\title{Adversarial Universal Stickers: Universal Perturbation Attacks on Traffic Sign using Stickers}

%%
%% Authors
\ifanonymousversion
\author{Anonymous Submission KDD 2025}

\else

\author{Anthony Etim}
\affiliation{%
  \institution{Yale University}
  \city{New Haven} 
  \state{Connecticut} 
  \country{USA}
}
\email{anthony.etim@yale.edu}

\author{Jakub Szefer}
\affiliation{%
  \institution{Northwestern University}
  \city{Evanston} 
  \state{IL} 
  \country{USA}
}
\email{jakub.szefer@northwestern.edu}

\fi

%%
%% Paper abstract
\begin{abstract}
    Adversarial attacks on deep learning models have proliferated in recent years. In many cases, a different adversarial perturbation is required to be added to each image to cause the deep learning model to misclassify it. This is ineffective as each image has to be modified in a different way. Meanwhile, research on universal perturbations focuses on designing a single perturbation that can be applied to all images in a data set, and cause a deep learning model to misclassify the images. This work advances the field of universal perturbations by exploring universal perturbations in the context of traffic signs and autonomous vehicle systems. This work introduces a novel method for generating universal perturbations that visually look like simple black and white stickers, and using them to cause incorrect street sign predictions. Unlike traditional adversarial perturbations, the adversarial universal stickers are designed to be applicable to any street sign: same sticker, or stickers, can be applied in same location to any street sign and cause it to be misclassified. Further, to enable safe experimentation with adversarial images and street signs, this work presents a virtual setting that leverages Street View images of street signs, rather than the need to physically modify street signs, to test the attacks. The experiments in the virtual setting demonstrate that these stickers can consistently mislead deep learning models used commonly in street sign recognition, and achieve high attack success rates on dataset of US traffic signs. The findings highlight the practical security risks posed by simple stickers applied to traffic signs, and the ease with which adversaries can generate adversarial universal stickers that can be applied to many street signs.
\end{abstract}

%%
%% Make title area
\maketitle

%%
%% Show page numbers, remove for final document
\pagestyle{plain}

%%
%% Main sections

\section{Introduction}
\label{sec_introduction}

Deep Neural Networks (DNNs) have revolutionized machine learning applications, yet they remain inherently vulnerable to adversarial perturbations -- subtle, often imperceptible modifications to input data that can drastically alter model predictions~\cite{chakraborty2018adversarial}. First identified by Szegedy et al.~\cite{szegedy2013intriguing}, these adversarial examples expose fundamental weaknesses in deep learning models, particularly in safety-critical domains like autonomous driving, biometric authentication, and medical imaging~\cite{goodfellow2014explaining},~\cite{kumar2024medical}. As DNNs continue to be integrated into real-world applications, their susceptibility to adversarial attacks raises significant concerns about security, reliability, and trustworthiness.

One of key domains where DNNs can be applied is autonomous vehicles, where street sign recognition can be easily done by the DNN to help an autonomous vehicle navigate physical world. Unfortunately, adversarial perturbations can prevent correct street sign recognition or classification. Existing adversarial perturbation work has been applied in the context of traffic sign recognition and object detection, and attackers have demonstrated various perturbation strategies. Existing attacks include physical sticker attacks~\cite{eykholt2018robust} where various stickers are placed on street signs to cause the street sign to not be correctly recognized. More recent work on adversarial shadows~\cite{zhong2022shadows} shows that light shadows can achieve same effect; the shadows are simple, however, generating them in practice may be difficult, e.g., on a cloudy day. Other work has shown light-based manipulations~\cite{hsiao2024natural} where bright light spots are shone on street signs to cause them to be misclassified. All of these  attacks leverage model vulnerabilities to induce erroneous street sign predictions. These attacks highlight not only the fragility of DNN-based perception systems but also the practical feasibility of adversarial manipulations in physical environments.

One limitation of the attacks is that the perturbations are specific to each traffic sign. In other setting, universal adversarial perturbations (UAPs) have been shown to pose a particularly severe threat, as they can fool a model across multiple inputs with a single, input-agnostic perturbation~\cite{moosavi2017universal}. Unlike targeted adversarial attacks, including the existing ones that attack street signs~\cite{eykholt2018robust,zhong2022shadows,hsiao2024natural},  UAPs generalize across diverse samples, significantly compromising model robustness in real-world scenarios~\cite{weng2024comparative}. 

In this work, we investigate the impact of universal adversarial perturbations (UAPs) on the specific case of traffic sign classification systems. We focus on the design of simple black and white stickers, and find how these stickers can be placed at a common coordinate on any street sign to cause it to be misclassified. Our experiments demonstrate that a single location can be exploited across multiple sign types to consistently induce misclassification, highlighting a critical vulnerability in deep learning-based recognition systems. Without a need for a complex perturbation, a simple black or white sticker can cause a street sign to be misclassified and have the machine learning model generate confidence score of up to 90\%. Further, unlike targeted attacks that require sign-specific modifications, our findings reveal that single location where one or two stickers are located on any traffic sign can cause misclassification, enabling a universal attack strategy that generalizes across different street signs.

Further, to enable safe experimentation with adversarial images and street signs, this work presents a virtual setting that leverages Street View images of street signs, rather than need to physically modify street signs, to test the attacks. Due to practical constraints, and considering potential safety issues, actual street signs were not used nor modified for this work. The attacks are evaluated by modifying images from Street View. This further demonstrates that attackers can leverage easily accessible Street View images to test their attacks and find the locations for the adversarial universal stickers to place on the street signs.

Our insights have significant implications for the security of autonomous driving systems, as the suggests that attackers can design efficient, low-cost perturbations capable of misleading a model without requiring precise adjustments for each individual sign. Additionally, our results emphasize the need for more robust defense mechanisms that account for spatially invariant attack strategies, as current adversarial training and input transformation techniques may fail against universal perturbations that exploit shared model weaknesses.

\subsection{Contributions}

The contributions of this work are as follows:

\begin{enumerate}

    \item We demonstrate the first example of universal perturbation attacks targeting specifically street signs.

    \item We show that it is possible to enable safe experimentation with adversarial images and street signs by using a virtual setting that leverages Street View images of real street signs, without need to actually modify real signs -- which may be illegal or dangerous.

    \item We demonstrate that a simple black or white sticker located strategically on the face of the street sign can cause misclassification with confidence score (of the incorrect classification) as high as 90\%.

    \item We show that for single or double white or black stickers it is possible to find common coordinates on the tested street sign such that each street sign is misclassified.

\end{enumerate}

\section{Background}
\label{background}

In this section, we provide a brief overview of adversarial attacks and the LISA dataset, which contains images of U.S. street signs. We also discuss LISA-CNN, the victim machine learning model used for street sign classification in this work. While we utilize LISA-CNN, we apply it to current Street View images of traffic signs.

\subsection{Adversarial Attacks}

Machine learning models, particularly deep neural networks (DNNs), are vulnerable to adversarial attacks -- small, carefully crafted modifications to input data that can alter predictions. These attacks operate in two primary settings: white-box, where the attacker has full access to the model’s parameters, and black-box, where the attacker has no direct knowledge of the target model~\cite{chakraborty2018adversarial,liang2022adversarial}.

One of the earliest attack methods, Fast Gradient Sign Method (FGSM), perturbs an image by adding the sign of the gradient of the loss function~\cite{goodfellow2014explaining}. More advanced iterative techniques, such as Projected Gradient Descent (PGD), refine this approach to generate stronger adversarial examples~\cite{mkadry2017towards}. While these attacks typically generate unique perturbations for each input, a more scalable approach is universal adversarial perturbations (UAPs)—a single perturbation designed to fool a model across multiple inputs~\cite{moosavi2017universal}.

Adversarial patches take this concept further by localizing perturbations to a specific region within an image rather than modifying the entire input~\cite{brown2017adversarial}. These patches are often designed to be universal, making them particularly practical for real-world attacks. To enhance their robustness against variations in viewing angles, lighting, and environmental factors, Expectation over Transformations (EoT) was introduced to generate patches that remain effective under common distortions~\cite{athalye2018synthesizing}. However, real-world studies, such as Robust Physical Perturbations (RP2), demonstrate that synthetic transformations alone fail to fully capture environmental complexity, necessitating real-world testing under diverse conditions~\cite{eykholt2018robust}.

Beyond noise-based and patch-based attacks, light and shadow-based adversarial attacks have emerged as more stealthy alternatives. Shadow attacks leverage natural or artificial shadows to create disruptions that are difficult for both human observers and AI models to detect~\cite{zhong2022shadows}. Similarly, light-based attacks manipulate natural light sources, such as sunlight or flashlights, to subtly distort object recognition~\cite{hsiao2024natural}. Leaf-based adversarial attacks leverage naturally occurring leaves to fool the machine learning model~\cite{etim2024fallleafadversarialattack}. These methods exploit everyday environmental factors, posing a significant challenge for AI-driven vision systems in real-world~settings.

\subsection{LISA-CNN and LISA Traffic Sign Dataset}

The LISA dataset is a collection of U.S. traffic sign images, covering 47 different road sign categories~\cite{lisa}. Due to class imbalances, prior research has focused on a subset of the $16$ most common signs, a selection that improves model training efficiency and classification performance~\cite{hsiao2024natural}.

LISA-CNN is a convolutional neural network designed specifically for traffic sign recognition using this dataset. The architecture consists of three convolutional layers followed by a fully connected layer, optimized to classify traffic signs under various environmental conditions and viewpoints~\cite{eykholt2018robust}. Models trained on LISA have demonstrated high accuracy and real-time detection capabilities, making them well-suited for autonomous vehicle perception~\cite{pavlitska2023adversarial} In this work, we utilize LISA-CNN to analyze adversarial universal perturbations on traffic signs. As this is the most common model used for traffic sign recognition, it is only one used in this work, but the ideas should apply equally to other models.

\subsection{Street View Street Sign Images}

Due to practical constraints, and considering potential safety issues, actual street signs are not used nor modified for this work. The attacks are evaluated by modifying images from Street View~\cite{googleExploreStreet}. Since the evaluation is done on real Street View images, we believe it emulates well real-world attacks on physical street signs without need to actually modify real street signs. Further, the use of Street View images ensures that the training set (LISA images) is different from the testing set (Street View images).

\begin{figure*}[t]
    \centering
    \includegraphics[width=15cm]{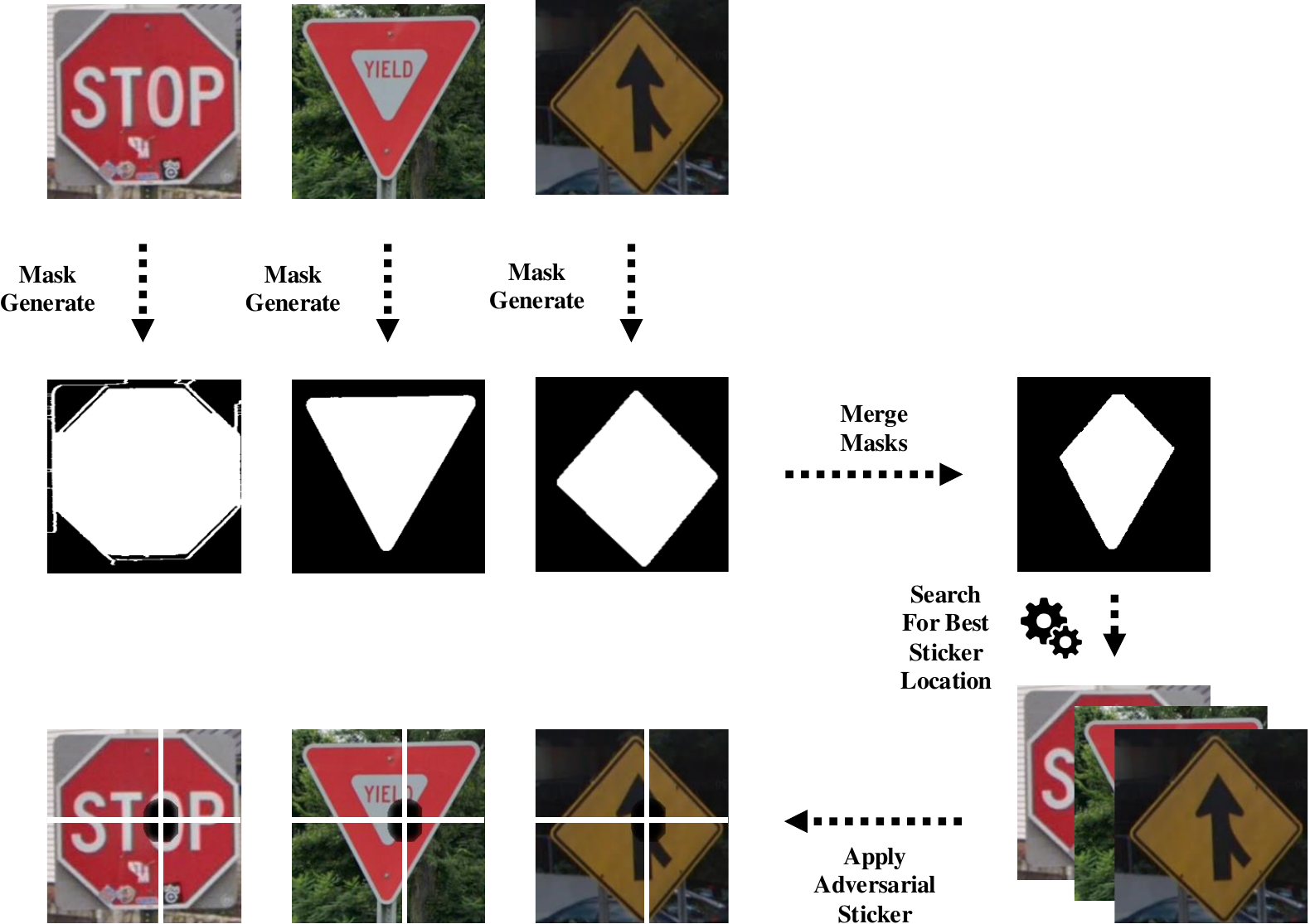}
    \caption{\small Adversarial universal sticker attack workflow.}
\label{fig_universal_perturbation_workflow}
\end{figure*}

\section{Threat Model}

In this work, we consider an adversary aiming to manipulate a deep learning-based traffic sign classification system using universal perturbations. The attacker has black-box access to the victim model, meaning they do not have direct knowledge of its architecture, parameters, or training data but can observe its predictions. The attack assumes that the adversary can place small, inconspicuous perturbations, such as stickers, on traffic signs in real-world environments to trigger the attack.

The primary objective of the attacker is to induce misclassification across multiple sign types using a single, common adversarial perturbation. This perturbation is designed to exploit shared vulnerabilities in the model’s feature extraction process, ensuring that the perturbation generalizes across different traffic signs.

We assume that the system relies solely on visual input for traffic sign recognition, without additional contextual verification mechanisms such as sensor fusion or map-based validation. Furthermore, we consider scenarios where standard adversarial defenses, such as adversarial training or input preprocessing, may be ineffective against universal physical perturbations. This threat model reflects real-world risks in autonomous driving and intelligent transportation systems, where adversarial manipulation of traffic signs could lead to incorrect or unsafe driving decisions.

\section{Universal Perturbation Attack on Street Signs}

This section introduces a universal perturbation attack targeting traffic sign classification systems. Unlike traditional per-instance adversarial attacks, our approach applies a common perturbation across multiple traffic signs to induce consistent misclassification. By strategically placing a universal adversarial mask in a fixed location, we demonstrate the vulnerability of deep learning models to input-agnostic physical perturbations. Our attack is evaluated on real-world traffic sign images, highlighting its robustness under varying environmental conditions. This study underscores the risks posed by universal perturbations and their potential impact on machine learning-driven traffic sign recognition systems.

\subsection{Attack Workflow}

Figure~\ref{fig_universal_perturbation_workflow} shows the attack workflow. The goal of this work is to design simple, yet effective framework for universal perturbations on street signs. In our context {\em universal perturbation means a perturbation that is applied to the same location on the face of any street sign.} As a first step to achieving this, the street signs need to be analyzed to generate the universal perturbation mask region. This is the region common to all street signs such that a perturbation, i.e. sticker, can be placed on all street signs in same location.

In order to find the common region, the street signs are analyzed to generate a mask, i.e. identify the region occupied by the street sign, as shown in middle-left of Figure~\ref{fig_universal_perturbation_workflow}. Next, the masks are merged, to generate the universal perturbation mask region, as shown in middle-right of Figure~\ref{fig_universal_perturbation_workflow}. Any sticker placed within the merged mask region is quarantined to fit on any of the street signs.

Next, the best location for the stickers is searched within the merged mask. An exhaustive search is performed where a sticker is placed at the same location on each street sign, and the confidence scores are collected. The location that gives the highest average confidence score (for a wrongly predicted street sign) is selected as the target location. Finally, the selected location is used to apply the sticker, as shown in bottom-left of the Figure~\ref{fig_universal_perturbation_workflow}. Note that the white cross-marks are only shown to illustrate and highlight that the stickers are placed in same location on all the street signs, they are not part of the attack.

\begin{figure*}[t]
    \begin{subfigure}[b]{0.25\textwidth}
        \centering
\includegraphics[width=2.2cm]{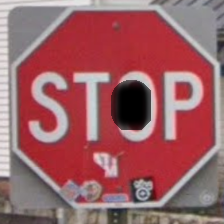}
        \caption{\small \centering Stop Black Sticker Image}
        \label{fig:Best_stop_black_one_sticker}
    \end{subfigure}
    \hfill
    \begin{subfigure}[b]{0.25\textwidth}
        \centering
         \includegraphics[width=2.2cm]{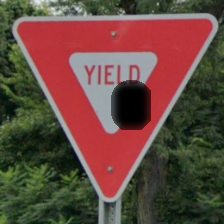}
        \caption{\small \centering Yield Black Sticker Image}
        \label{fig:Best_yield_black_one_sticker}
    \end{subfigure}
    \hfill
    \begin{subfigure}[b]{0.25\textwidth}
        \centering
   \includegraphics[width=2.2cm]{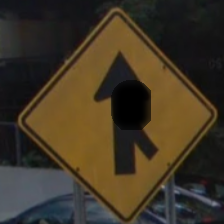}
        \caption{\small \centering Merge Black Sticker Image}
        \label{fig:Best_merge_black_one_sticker}
    \end{subfigure}
    \vspace{-1em}
    \caption{Single Black sticker attack images.}
    \label{fig:one_sticker_black_Images}
\end{figure*}

\begin{figure*}[t]
    \begin{subfigure}[b]{0.25\textwidth}
        \centering
\includegraphics[width=2.2cm]{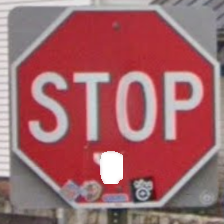}
        \caption{\small \centering Stop White Sticker Image}
        \label{fig:Best_stop_white_one_sticker}
    \end{subfigure}
    \hfill
    \begin{subfigure}[b]{0.25\textwidth}
        \centering
         \includegraphics[width=2.2cm]{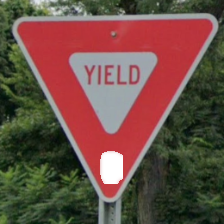}
        \caption{\small \centering Yield White Sticker Image}
        \label{fig:Best_yield_white_one_sticker}
    \end{subfigure}
    \hfill
    \begin{subfigure}[b]{0.25\textwidth}
        \centering
   \includegraphics[width=2.2cm]{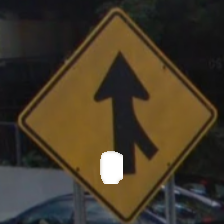}
        \caption{\small \centering Merge White Sticker Image}
        \label{fig:Best_merge_white_one_sticker}
    \end{subfigure}
    \vspace{-1em}
    \caption{Single White sticker attack images.}
    \label{fig:one_sticker_white_Images}
\end{figure*}

The adversarial stickers used in this work are simple black or white stickers. Unlike existing work on (non-universal) adversarial stickers~\cite{9779913} which used various colored stickers, our work focuses on simple stickers with just two colors. Examples of stickers, overlaid on top of street signs, are shown later, e.g., in Figures~\ref{fig:one_sticker_black_Images} to~\ref{fig:two_sticker_white_white_Images}.

\section{Experimental Results}
\label{experimental_results}

This section presents the evaluation of the adversarial universal stickers attack.

\subsection{Baseline Performance of Street Sign Classification}

The evaluation is performed on 3 randomly selected images: Stop, Yield, and Merge. The input images are from Street View, not from the LISA data set. Thus the training set (LISA images) is different from the testing set (Street View images). LISA-CNN performs well, with average over $80$\% confidence scores in the correctly predicted images. Specifically, Table~\ref{table_adversarial} shows the confidence scores for the street signs.

\begin{table}[t]
\centering
\caption{Baseline images and their confidence scores.}
%\begin{adjustbox}{width=0.48\textwidth}
\label{table_adversarial}
\small
\begin{tabular}{|p{1.8cm}|p{1.8cm}|p{1.8cm}|p{1.2cm}|}
\hline 
\textbf{Image} & \textbf{Predicted Label}  & \textbf{Confidence Score (\%)}
\\ \hline \hline
Stop &  Stop &  85.42 \\ \hline 
Yield  & Yield & 88.87 \\ \hline 
Merge & Merge & 76.51 \\ \hline 

\end{tabular}
%\end{adjustbox}
\end{table}

\subsection{One Sticker Attacks}

The first evaluation was done by generating black and white stickers, one at a time. Stickers of rectangular shape were generated. The width and height were selected to be in ranges from 5\% to 50\% of the street sign size. All combinations of widths and heights form 5 to 50, in step of 5, were tested. Figures~\ref{fig:one_sticker_black_Images} and~\ref{fig:one_sticker_white_Images} show the best location for the black and white stickers respectively. Table~\ref{table_adversarial_timing} shows the confidence scores, and Table~\ref{table_adversarial_labels} shows the (incorrectly classified) labels for these street signs. It can be seen that with one white or black sticker, placed at same location or any of the signs, the machine learning classification results are highly incorrect and attack can succeed.

\begin{figure*}[t]
    \begin{subfigure}[b]{0.25\textwidth}
        \centering
\includegraphics[width=2.2cm]{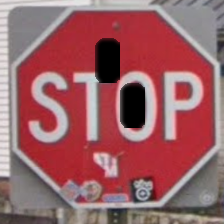}
        \caption{\small \centering Stop Black Stickers Image}
        \label{fig:Best_stop_black_black_two_sticker}
    \end{subfigure}
    \hfill
    \begin{subfigure}[b]{0.25\textwidth}
        \centering
         \includegraphics[width=2.2cm]{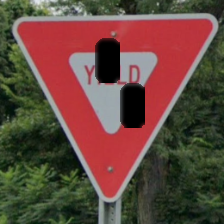}
        \caption{\small \centering Yield Black Stickers Image}
        \label{fig:Best_yield_black_black_two_sticker}
    \end{subfigure}
    \hfill
    \begin{subfigure}[b]{0.25\textwidth}
        \centering
  \includegraphics[width=2.2cm]{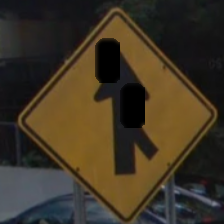}
        \caption{\small \centering Merge Black Stickers Image}
        \label{fig:Best_merge_black_black_two_sticker}
    \end{subfigure}
    \vspace{-1em}
    \caption{Two Black sticker attack images.}
    \label{fig:two_sticker_black_black_Images}
\end{figure*}

\begin{figure*}[t]
    \begin{subfigure}[b]{0.25\textwidth}
        \centering
\includegraphics[width=2.2cm]{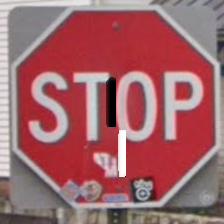}
        \caption{\small \centering Stop Black and White Stickers Image}
        \label{fig:Best_stop_black_white_two_sticker}
    \end{subfigure}
    \hfill
    \begin{subfigure}[b]{0.25\textwidth}
        \centering
         \includegraphics[width=2.2cm]{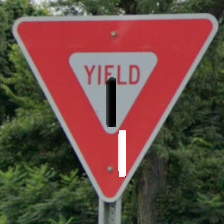}
        \caption{\small \centering Yield Black and White Stickers Image}
        \label{fig:Best_yield_black_white_two_sticker}
    \end{subfigure}
    \hfill
    \begin{subfigure}[b]{0.25\textwidth}
        \centering
  \includegraphics[width=2.2cm]{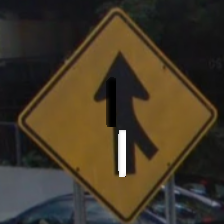}
        \caption{\small \centering Merge Black and White Stickers Image}
        \label{fig:Best_merge_black_white_two_sticker}
    \end{subfigure}
    \vspace{-1em}
    \caption{Two Black and White sticker attack images.}
    \label{fig:two_sticker_black_white_Images}
\end{figure*}

\begin{figure*}[t]
    \begin{subfigure}[b]{0.25\textwidth}
        \centering
\includegraphics[width=2.2cm]{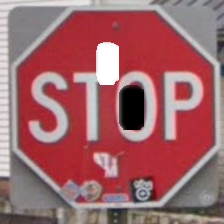}
        \caption{\small \centering Stop White and Black Stickers Image}
        \label{fig:Best_stop_white_black_two_sticker}
    \end{subfigure}
    \hfill
    \begin{subfigure}[b]{0.25\textwidth}
        \centering
         \includegraphics[width=2.2cm]{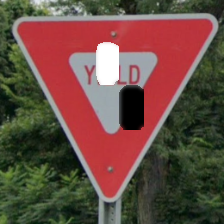}
        \caption{\small \centering Yield White and Black Stickers Image}
        \label{fig:Best_yield_white_black_two_sticker}
    \end{subfigure}
    \hfill
    \begin{subfigure}[b]{0.25\textwidth}
        \centering
  \includegraphics[width=2.2cm]{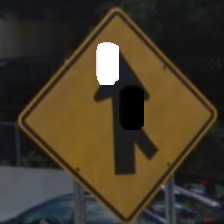}
        \caption{\small \centering Merge White and Black Stickers Image}
        \label{fig:Best_merge_white_black_two_sticker}
    \end{subfigure}
    \vspace{-1em}
    \caption{Two White and Black sticker attack images.}
    \label{fig:two_sticker_white_black_Images}
\end{figure*}

\begin{figure*}[h!]
    \begin{subfigure}[b]{0.25\textwidth}
        \centering
\includegraphics[width=2.2cm]{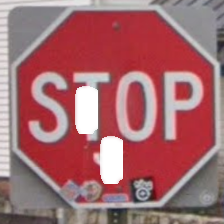}
        \caption{\small \centering Stop White and White Stickers Image}
        \label{fig:Best_stop_white_white_two_sticker}
    \end{subfigure}
    \hfill
    \begin{subfigure}[b]{0.25\textwidth}
        \centering
         \includegraphics[width=2.2cm]{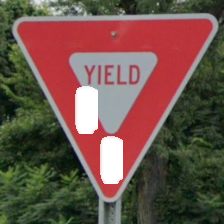}
        \caption{\small \centering Yield White and White Stickers Image}
        \label{fig:Best_yield_white_white_two_sticker}
    \end{subfigure}
    \hfill
    \begin{subfigure}[b]{0.25\textwidth}
        \centering
 \includegraphics[width=2.2cm]{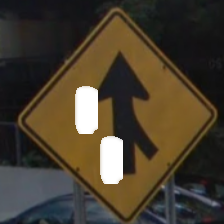}
        \caption{\small \centering Merge White and White Stickers Image}
        \label{fig:Best_merge_white_white_two_sticker}
    \end{subfigure}
    \vspace{-1em}
    \caption{Two White and White sticker attack images.}
    \label{fig:two_sticker_white_white_Images}
\end{figure*}

\subsection{Two Stickers}

The second evaluation was done by generating two black and white stickers, two at a time. Stickers of rectangular shape were again generated. The width and height were selected to be in ranges from 5\% to 50\% of the street sign size. All combinations of widths and heights form 5 to 50, in step of 5, were tested. Figures~\ref{fig:two_sticker_black_black_Images} to~\ref{fig:two_sticker_white_white_Images} show the best location for the different combinations of black and white stickers. Table~\ref{table_adversarial_timing} shows the confidence scores, and Table~\ref{table_adversarial_labels} shows the (incorrectly classified) labels for these street signs. It can be seen that with two white or black stickers in any combination, placed at same location or any of the signs, the machine learning classification results are highly incorrect and attack can succeed.

\subsection{Confidence Scores and Misclassified Labels Patterns}

From Table~\ref{table_adversarial_timing} and Table~\ref{table_adversarial_labels} we observed two patterns.
First, almost always it is possible to find a universal adversarial sticker for one or two stickers. The range of the confidence values (for the incorrectly classified signs) ranges from about 23\% to over 95\%. Only in two cases one of the signs was not misclassified.
Second, when misclassification occurs, the image is most often misclassified as the pedestrian crossing sign. As we treat LISA-CNN as a black-box, we do not have insights into why the pedestrian crossing sign shows up most often. However, this could be abused in future attacks where the attacker knows that the misclassification is likely to give a certain street sign class. E.g., instead of merging on a highway, the vehicle will (incorrectly) detect pedestrian crossing sign and stop, resulting in a crash on a highway.

\subsection{Sticker Size Evaluation}

We performed further extensive study to test what sticker sizes work best. Again, all combinations of widths and heights form 5 to 50, in step of 5, were tested. Tables~\ref{table_one_sticker_black} to~\ref{table_two_sticker_first_white_second_white} show the results.
For single black sticker, highest average confidence in correctly classified sign was over 80\% for 40x50 sticker.
For single white sticker, highest average confidence in correctly classified sign was over 42\% for 25x35 sticker.
For two stickers, black and black, highest average confidence in correctly classified sign was over 63\% for 25x45 sticker.
For two stickers, black and white, highest average confidence in correctly classified sign was over 56\% for 10x50 sticker.
For two stickers, white and black, highest average confidence in correctly classified sign was over 70\% for 25x45 sticker.
And for two stickers, white and white, highest average confidence in correctly classified sign was over 39\% for 25x50 sticker.

\subsection{Sticker Size Patterns}

We observed some expected and some unexpected pattern. For single black sticker, as the sticker size increases (form top-left to bottom-right in the table), the confidence in misclassified images increases. This follows the intuition that as the sticker gets bigger, it obscures larger portion of the street sign, making it less likely to be recognized correctly. On the other hand, for single white sticker, this pattern does not hold. Going form top-left to bottom-right in the table, as sticker dimensions increase, the confidence increases, but then half-way through the table it starts to decrease.

For two stickers, we observed a yet different, but consistent pattern. Going form top-left to bottom-right in the table, as sticker dimensions increase, the confidence in misclassified signs increases, then decreases a bit, until finally the size of the combined stickers is too large and does not fit in the mask area (X in the table entries). This pattern also makes sense. The intuition is that as stickers get bigger, they cause the signs to be incorrectly classified, similar to one black sticker. However, due to size of the stickers, no universal sticker can be found since for large sizes two stickers do not fit in the combined mask.

\begin{table*}[t]
\centering
\caption{Confidence scores of the best universal adversarial images.}
% \begin{adjustbox}{width=0.98\textwidth}
\label{table_adversarial_timing}
\small
\begin{tabular}{|p{1.5cm}|p{1.5cm}|p{1.5cm}|p{1.5cm}|p{1.5cm}|p{1.5cm}|p{1.5cm}|}
\hline 
\textbf{Adversarial Image} & \textbf{One Sticker Black}  & \textbf{One Sticker  White} & \textbf{Two Sticker Black, Black} & \textbf{Two Sticker Black, White} & \textbf{Two Sticker White, Black} & \textbf{Two Sticker White, White} 
\\ \hline \hline
Stop & 86.85 & 37.79 & 60.67 & 49.17& 67.25 & 25.98   \\ \hline 
Yield  & 65.44 & $\times$ & 55.24  & 23.70& 58.87& $\times$ \\ \hline 
Merge & 88.42 & 90.49 & 75.27 &95.41  &87.05 &91.50  \\ \hline 
\end{tabular}
% \end{adjustbox}
\end{table*}

\begin{table*}[t]
\centering
\caption{Predicted labels of the best universal adversarial images.}
% \begin{adjustbox}{width=0.98\textwidth}
\label{table_adversarial_labels}
\small
\begin{tabular}{|p{1.5cm}|p{1.5cm}|p{1.5cm}|p{1.5cm}|p{1.5cm}|p{1.5cm}|p{1.5cm}|}
\hline 
\textbf{Adversarial Image} & \textbf{One Sticker Black}  & \textbf{One Sticker  White} & \textbf{Two Sticker Black, Black} & \textbf{Two Sticker Black, White} & \textbf{Two Sticker White, Black} & \textbf{Two Sticker White, White} 
\\ \hline \hline
Stop & Ped. Crossing & Speed Limit 25& Ped. Crossing & Ped. Crossing& Ped. Crossing & Ped. Crossing   \\ \hline 
Yield  & Ped. Crossing & Yield & Ped. Crossing  & Ped. Crossing& Ped. Crossing& Yield \\ \hline 
Merge & Ped. Crossing & Ped. Crossing &Ped. Crossing&Ped. Crossing  &Ped. Crossing&Ped. Crossing \\ \hline 
\end{tabular}
% \end{adjustbox}
\end{table*}

\begin{table*}[t]
    \centering
    \caption{Average Confidence scores for universal adversarial images using a single black sticker.}
    \label{table_one_sticker_black}
    \begin{adjustbox}{width=\textwidth}
    \small
        \begin{tabular}{|p{1.5cm}|p{1.5cm}|p{1.5cm}|p{1.5cm}|p{1.5cm}|p{1.5cm}|p{1.5cm}|p{1.5cm}|p{1.5cm}|p{1.5cm}|p{1.5cm}|p{1.5cm}|}
            \hline
            \textbf{Height, Width} & 
            \textbf{5} &
            \textbf{10} & \textbf{15} & \textbf{20} & \textbf{25} & \textbf{30} & \textbf{35} & \textbf{40} & \textbf{45} & \textbf{50} \\ \hline
             \textbf{5} & - & - & - & - & - & - & - & - & - & -  \\ \hline
             \textbf{10} & - & - & - & - & 12.83 & 14.63 & 16.31 & 19.19 & 21.05  & 31.10 \\ \hline
             \textbf{15} & - & - & - & 15.74 & 19.39 & 22.46 & 26.04 & 36.27 & 41.86 & 47.98 \\ \hline
             \textbf{20} & - & - & 15.62 & 20.42 & 23.79 & 26.17 & 32.09 & 43.09 & 49.95 & 54.09 \\ \hline
            \textbf{25} & - & 14.79 & 19.28 & 24.75 & 26.85 & 30.43 & 42.95 & 55.94 & 62.54 & 72.31 \\ \hline
            \textbf{30} & - & 16.57 & 23.03 & 26.98 & 31.69 & 27.27 & 40.36 & 53.55 & 60.12 & 66.01 \\ \hline
           \textbf{35} & - & 18.07 & 24.91 & 28.01 & 45.13 & 40.38 & 47.83 & 59.38 & 67.27 & 73.54\\ \hline
            \textbf{40} & - & 19.63 & 26.07 & 35.19 & 53.13 & 49.37 & 56.81 & 65.90 & 73.55 & \textbf{80.24}  \\ \hline
           \textbf{45} & - & 20.69 & 33.72 & 41.32 & 55.24 & 54.27 & 61.89 & 69.23 & 70.08 & 77.52 \\ \hline
            \textbf{50} & - & 21.11 & 38.51 & 43.98 & 57.20 & 57.76 & 64.63 & 72.11 & 73.72 & 80.23 \\ \hline
        \end{tabular}
    \end{adjustbox}
\end{table*}

\begin{table*}[t]
    \centering
    \caption{Average Confidence scores for universal adversarial images using a single white sticker.}
    \label{table_one_sticker_white}
    \begin{adjustbox}{width=\textwidth}
    \small
        \begin{tabular}{|p{1.5cm}|p{1.5cm}|p{1.5cm}|p{1.5cm}|p{1.5cm}|p{1.5cm}|p{1.5cm}|p{1.5cm}|p{1.5cm}|p{1.5cm}|p{1.5cm}|p{1.5cm}|}
            \hline
            \textbf{Height, Width} & 
            \textbf{5} &
            \textbf{10} & \textbf{15} & \textbf{20} & \textbf{25} & \textbf{30} & \textbf{35} & \textbf{40} & \textbf{45} & \textbf{50} \\ \hline
             \textbf{5}  & -  & -  & 20.25  &  29.53  & 30.60  & 31.41  & 31.79  & 31.81  & 31.80  & 31.36  \\ \hline
            \textbf{10}  & -  & -  & 23.51  & 31.68  & 31.98  & 32.01  & 31.98  & 31.83  & 31.40  & 31.35  \\ \hline
            \textbf{15}  & 20.28  & 30.07  & 31.86  & 31.98  &  31.98 & 31.79  & 31.56  & 31.43  & 31.32  & 29.99  \\ \hline
            \textbf{20}  & 29.49 & 31.74  & 32.03  & 31.98  & 31.85  & 31.56  & 31.15  &  31.03 & 30.57  & 30.53  \\ \hline
            \textbf{25}  & 30.53  & 32.01  & 32.04 & 31.67  & 31.67  & 41.72  & \textbf{42.76}  & 31.52  & 31.26  & 30.47  \\ \hline
            \textbf{30}  & 31.35  & 31.84  & 31.80  & 37.95  & 40.34  & 31.41  & 31.56  & 31.52  & 31.18 & 27.62  \\ \hline
            \textbf{35}  & 31.50  & 31.78  & 31.68 &  32.13 & 32.14 & 31.65 & 31.77  & 31.31 & 28.21  & 25.42  \\ \hline
            \textbf{40}  & 31.54  & 31.70  & 31.73  & 39.55  & 40.35 & 31.83  &  30.79  & 30.03  & 24.59  & 22.88  \\ \hline
            \textbf{45}  & 31.35 & 31.77  & 31.79  &  32.41 & 32.38  & 28.75  & 25.55 & 23.86  & 21.17  & 18.85  \\ \hline
            \textbf{50}  & 31.33  & 31.91  & 31.96  & 31.73  & 27.75  & 24.33  & 23.85  & 19.96  & 19.51  & 18.59  \\ \hline
        \end{tabular}
    \end{adjustbox}
\end{table*}

\begin{table*}[t]
    \centering
    \caption{Average Confidence scores for universal adversarial images using two stickers (first black, second black).\vspace{-0.5em}}
\label{table_two_sticker_first_black_second_black}
    \begin{adjustbox}{width=\textwidth}
    \small
        \begin{tabular}{|p{1.5cm}|p{1.5cm}|p{1.5cm}|p{1.5cm}|p{1.5cm}|p{1.5cm}|p{1.5cm}|p{1.5cm}|p{1.5cm}|p{1.5cm}|p{1.5cm}|}
            \hline
            \textbf{Height, Width} & 
            \textbf{5} &
            \textbf{10} & \textbf{15} & \textbf{20} & \textbf{25} & \textbf{30} & \textbf{35} & \textbf{40} & \textbf{45} & \textbf{50} \\ \hline
             \textbf{5}  & -  & -  & -  & -  & -  & -  & -  & -  & -  & -  \\ \hline
            \textbf{10}  & -  & -  & -  & 11.40  & 14.41  & 16.22  & 18.67  & 27.16  & 31.69  & 34.01  \\ \hline
            \textbf{15}  & -  & -  & 12.74  & 17.13  & 25.00  & 31.09  & 33.73  & 36.89  & 40.49  & 43.55  \\ \hline
            \textbf{20}  & -  & 12.66  & 17.20  & 29.86  & 35.86  & 42.68  & 48.24  & 49.70  & 52.02  & 53.02  \\ \hline
            \textbf{25}  & -  & 16.06  & 27.41  & 34.65  & 42.22  & 50.39  & 53.44  & 57.99  & \textbf{63.73}  & 55.19  \\ \hline
            \textbf{30}  & -  & 17.14  & 28.85  & 40.52  & 44.21  & 34.50  & 38.14  & 42.35  & 39.08  & 20.87  \\ \hline
            \textbf{35}  & -  & 18.45  & 32.36  & 38.02  & 44.16  & 39.94  & 36.15  & 31.07  & 11.03  & $\times$  \\ \hline
            \textbf{40}  & -  & 17.11  & 41.25  & 41.75  & 39.77  & 31.83  & 29.96  & $\times$  & $\times$  & $\times$  \\ \hline
            \textbf{45}  & -  & 16.75  & 33.42  & 40.16  & 33.63  & 7.80  & $\times$  & $\times$ & $\times$  & $\times$ \\ \hline
            \textbf{50}  & -  & 17.15  & 27.93  & 38.82  & $\times$  & $\times$ & $\times$  & $\times$ & $\times$  & $\times$ \\ \hline
        \end{tabular}
    \end{adjustbox}
\end{table*}

\begin{table*}[t]
    \centering
     \caption{Average Confidence scores for universal adversarial images using two stickers (first black, second white).}
    \label{table_two_sticker_first_black_second_white}
    \begin{adjustbox}{width=\textwidth}
    \small
        \begin{tabular}{|p{1.5cm}|p{1.5cm}|p{1.5cm}|p{1.5cm}|p{1.5cm}|p{1.5cm}|p{1.5cm}|p{1.5cm}|p{1.5cm}|p{1.5cm}|p{1.5cm}|}
            \hline
            \textbf{Width, Height} & 
            \textbf{5} &
            \textbf{10} & \textbf{15} & \textbf{20} & \textbf{25} & \textbf{30} & \textbf{35} & \textbf{40} & \textbf{45} & \textbf{50} \\ \hline
             \textbf{5}  & -  & -          & 19.11  & 29.39  & 30.83  & 31.63  & 32.02  & 32.04  & 31.97  & 31.65  \\ \hline
            \textbf{10}  & -  & -  & 23.73          & 31.86  & 32.18  & 32.24  & 37.62  & 47.71  & 55.40  & \textbf{56.09}  \\ \hline
            \textbf{15}  & 16.59  & 29.42  & 31.94  & 32.23  & 43.31  & 46.49  & 44.18  & 53.84  & 48.54  & 46.16  \\ \hline
            \textbf{20}  & 29.40  & 31.73  & 32.06  & 32.15  & 44.74  & 44.30  & 43.09  & 42.43  & 43.47  & 36.32  \\ \hline
            \textbf{25}  & 31.19  & 31.88  & 31.72  & 39.44  & 41.62  & 39.19  & 40.92  & 36.12  & 31.94  & 36.66  \\ \hline
            \textbf{30}  & 31.76  & 31.97   & 31.55  & 34.09  & 31.23  & 30.22  & 30.29  & 29.22  & 28.30  & 32.04  \\ \hline
            \textbf{35}  & 31.65  & 31.55 & 29.34  & 28.05  & 32.18  & 28.12  & 25.71  & 24.54  & 17.05  & $\times$  \\ \hline
            \textbf{40}  & 24.96  & 25.32  & 20.39  & 26.00  & 28.17  & 22.81  & 20.06  & $\times$  & $\times$  & $\times$  \\ \hline
            \textbf{45}  & 20.61  & 20.85  & 20.25  & 22.68  & 22.46  & 15.51  & $\times$  & $\times$ & $\times$  & $\times$  \\ \hline
            \textbf{50}  & 19.09  & 19.43  & 16.56  & 18.23  & $\times$  & $\times$  & $\times$ & $\times$  & $\times$  & $\times$  \\ \hline
        \end{tabular}
    \end{adjustbox}
\end{table*}

\begin{table*}[t]
    \centering
    \caption{Average Confidence scores for universal adversarial images using two stickers (first white, second black).}
    \label{table_two_sticker_first_white_second_black}
    \begin{adjustbox}{width=\textwidth}
    \small
        \begin{tabular}{|p{1.5cm}|p{1.5cm}|p{1.5cm}|p{1.5cm}|p{1.5cm}|p{1.5cm}|p{1.5cm}|p{1.5cm}|p{1.5cm}|p{1.5cm}|p{1.5cm}|p{1.5cm}|}
            \hline
            \textbf{Height, Width} & 
            \textbf{5} &
            \textbf{10} & \textbf{15} & \textbf{20} & \textbf{25} & \textbf{30} & \textbf{35} & \textbf{40} & \textbf{45} & \textbf{50} \\ \hline
             \textbf{5}     &      - &      - &  20.25 &  27.59 &  27.79 &  24.40 &  24.83 &  25.08 &  25.90 &  24.97 \\ \hline
\textbf{10}    &      - &      - &  19.35 &  28.09 &  23.58 &  23.84 &  25.01 &  24.88 &  22.58 &  32.95 \\ \hline
\textbf{15}    &  18.65 &  19.23 &  22.63 &  23.44 &  23.00 &  22.35 &  25.45 &  35.11 &  45.80 &  52.51 \\ \hline
\textbf{20}    &  14.99 &  21.98 &  22.46 &  21.66 &  34.01 &  47.95 &  58.98 &  64.28 &  65.50 &  69.16 \\ \hline
\textbf{25}    &  17.56 &  22.78 &  20.23 &  34.17 &  49.53 &  61.45 &  66.84 &  70.02 &  \textbf{70.39} &  67.56 \\ \hline
\textbf{30}    &  18.89 &  19.09 &  28.26 &  47.14 &  60.37 &  54.23 &  56.72 &  60.39 &  59.07 &  41.45 \\ \hline
\textbf{35}    &  13.76 &  22.06 &  36.16 &  56.99 &  65.09 &  57.93 &  57.57 &  62.80 &  50.05 &      $\times$ \\ \hline
\textbf{40}    &  14.07 &  24.37 &  39.47 &  59.52 &  64.66 &  55.28 &  60.09 &      $\times$ &      $\times$ &      $\times$ \\ \hline
\textbf{45}    &  15.85 &  27.32 &  42.01 &  57.20 &  58.19 &  55.44 &     $\times$ &      $\times$ &     $\times$ &      $\times$ \\ \hline
\textbf{50}    &  18.06 &  27.92 &  41.33 &  41.64 &      $\times$ &      $\times$ &      $\times$ &      $\times$ &    $\times$  & $\times$ \\ \hline
        \end{tabular}
    \end{adjustbox}
\end{table*}

\begin{table*}[t]
    \centering
    \caption{Average Confidence scores for universal adversarial images using two stickers (first white, second white).}
    \label{table_two_sticker_first_white_second_white}
    \begin{adjustbox}{width=\textwidth}
    \small
        \begin{tabular}{|p{1.5cm}|p{1.5cm}|p{1.5cm}|p{1.5cm}|p{1.5cm}|p{1.5cm}|p{1.5cm}|p{1.5cm}|p{1.5cm}|p{1.5cm}|p{1.5cm}|p{1.5cm}|}
            \hline
            \textbf{Height, Width} & 
            \textbf{5} &
            \textbf{10} & \textbf{15} & \textbf{20} & \textbf{25} & \textbf{30} & \textbf{35} & \textbf{40} & \textbf{45} & \textbf{50} \\ \hline
            \textbf{5}  &      - &      - &  20.25 &  29.68 &  29.86 &  31.11 &  31.63 &  31.71 &  31.39 &  31.36 \\ \hline
\textbf{10} &      - &  21.04 &  31.39 &  30.85 &  29.81 &  28.94 &  28.87 &  28.94 &  28.50 &  27.01 \\ \hline
\textbf{15} &  24.50 &  31.33 &  31.27 &  31.32 &  30.50 &  28.84 &  28.30 &  27.67 &  28.01 &  29.08 \\ \hline
\textbf{20} &  29.39 &  31.31 &  31.21 &  30.29 &  28.29 &  26.81 &  26.21 &  27.30 &  28.70 &  29.88 \\ \hline
\textbf{25} &  29.81 &  31.24 &  31.37 &  30.77 &  30.04 &  28.93 &  29.07 &  29.51 &  30.19 &  \textbf{39.16} \\ \hline
\textbf{30} &  31.04 &  31.43 &  30.92 &  28.73 &  27.31 &  29.05 &  29.23 &  28.88 &  29.73 &  30.94 \\ \hline
\textbf{35} &  30.51 &  25.26 &  26.34 &  21.57 &  22.40 &  23.39 &  24.67 &  21.71 &  22.82 &      $\times$ \\ \hline
\textbf{40} &  21.59 &  21.11 &  23.39 &  22.56 &  23.91 &  24.00 &  23.81 &      $\times$ &      $\times$ &      $\times$ \\ \hline
\textbf{45} &  20.00 &  20.23 &  23.39 &  21.52 &  22.37 &  22.53 &      $\times$&      $\times$ &      $\times$ &      $\times$ \\ \hline
\textbf{50} &  17.90 &  20.94 &  21.81 &  18.03 &      $\times$ &      $\times$ &      $\times$ &      $\times$ &      $\times$&      $\times$ \\ \hline
            
        \end{tabular}
    \end{adjustbox}
\end{table*}

\section{Related Work}
\label{sec:related_work}

This section provides an overview of existing work on universal perturbation attacks as well as attacks using adversarial stickers.

\subsection{Universal Perturbation Attacks}

Existing universal adversarial attacks can be broadly grouped within two categories: noise-based 
and the generator-based~\cite{weng2024comparative}.
Previous noise-based Adversarial Universal attacks overlooked the link between instance-level and universal-level attacks. To address this, Li et al.~\cite{Li_Yang_Wei_Yang_Huang_2022} introduced AE-UAP, which learns universal adversarial perturbations by incorporating the instance-level adversarial examples into the learning to learn more dominant perturbations~\cite{weng2024comparative}. Furthermore, the authors in ~\cite{Weng_Luo_Zhong_Lin_Li_2023} tackled the dominant bias in ensemble attacks using a non-target Kullback–Leibler (KL) loss. Additionally, the min-max optimization method employed dynamically adjusts ensemble weights, further boosting attack transferability across different models.

Regarding the generator-based category, building on the Transferable Targeted Perturbations (TTP) framework~\cite{naseer2021generating},Wang et al. introduced the Transferable Targeted Adversarial Attack (TTAA) method~\cite{10204634}. Unlike TTP, which relies on neighborhood similarity matching, TTAA incorporates a feature discriminator to align the feature distributions of adversarial samples with those of the real target class. 
In addition, the authors in~\cite{10205119} built on the TTP framework with Minimizing Maximum Model Discrepancy (M3D), a method that enhances the transferability of black-box targeted attacks. They demonstrated that the generalization attack error on a black-box target model primarily depends on the empirical attack error on the substitute model and the maximum model discrepancy among substitute models.

In contrast, in this work we focus on simple, yet effective approach where the perturbation is achieved by adding sticker(s) on the face of the street signs. Simple black stickers can be easily and practically deployed in practice to modify the classification of the~signs.

\subsection{Adversarial Stickers}

Prior work has explored adversarial stickers, but not in a universal perturbation setting. Prior work such as Robust Physical Perturbation (RP2)~\cite{eykholt2018robust} has exposed the vulnerability of deep neural networks (DNNs) to adversarial examples in real-world settings. Designed for safety-critical applications, RP2 generates robust visual perturbations that mislead classifiers under various conditions. Using road sign classification as a case study, it demonstrated high success rates in both lab and field tests, causing targeted misclassification with simple black-and-white sticker perturbations.

The authors in ~\cite{9779913} propsed the Meaningful Adversarial Sticker, a novel, physically feasible, and stealthy attack method for black-box attacks in real-world settings. This approach manipulates everyday stickers by adjusting their placement, rotation, and other parameters to deceive computer vision systems. To optimize these parameters efficiently, the authors propose the  Region based Heuristic Differential Evolution (RHDE) algorithm, which incorporates an offspring generation strategy that aggregates effective solutions and an adaptive adjustment mechanism for evaluation criteria. Extensive experiments in both digital and physical environments—including face recognition, image retrieval, and traffic sign recognition—validated the method’s effectiveness. Remarkably, even without prior knowledge of the target model, the attack successfully misleads vision systems in a covert manner, exposing potential security vulnerabilities.

In contrast, this work demonstrated how simple stickers can be placed on street signs, each sticker in the same location on all the street signs. This removes a need for street-sign specific perturbations.

\section{Conclusion}
\label{sec_conclusion}

This work introduced a novel method for generating universal perturbations that visually look like simple black and white stickers, and using them to cause incorrect street sign prediction. Unlike traditional adversarial perturbations, the adversarial universal stickers are designed to be applicable to any street sign: same sticker, or stickers, can be applied in same location to any street sign and cause it to be misclassified. Further, to enable safe experimentation with adversarial images and street signs, this work presented a virtual setting that leverages Street View images of street signs, rather than need to physically modify street signs, to test the attacks. The experiments in the virtual setting demonstrated that these stickers can consistently mislead deep learning model used commonly in street sign recognition, and achieve high attack success rates on dataset of US traffic signs. The findings highlight the practical security risks posed by simple stickers applied to traffic signs, and the ease with which adversaries can generate adversarial universal stickers that can be applied to many street signs.

\section*{Acknowledgements}

This work was supported in part by National Science Foundation grant \nsf{2245344}.

%%
%% Bibliography
\bibliographystyle{ACM-Reference-Format}
\bibliography{bibtex/references}

%%
%% That's All Folks!
\end{document}